\documentclass[final,5p,times,nonatbib]{elsarticle}

\usepackage[T1]{fontenc}
\usepackage[utf8]{inputenc}
\usepackage{textcomp}
\usepackage{microtype}
\usepackage{graphicx}
\usepackage{amsmath}
\usepackage{booktabs}
\usepackage{array}
\usepackage{tabularx}
\usepackage{csquotes}
\usepackage[backend=biber,style=numeric-comp,sorting=none,natbib=true]{biblatex}
\makeatletter
\renewcommand*{\abx@missing}[1]{\mbox{\bfseries ??}}
\makeatother
\usepackage{url}
\usepackage{xurl}
\usepackage{newunicodechar}
\newunicodechar{−}{-}
\usepackage[hidelinks]{hyperref}

\journal{Smart Agricultural Technology}

\begin{document}

\begin{frontmatter}

\title{Event-Triggered Visual Perception for Real-Time Harvest Outcome Estimation in Mechanical Blueberry Harvesting}

\author[aff1,aff2]{Nicolas Miroslav Husár\corref{cor1}}
\author[aff2]{Jelle Driessen\corref{cor1}}
\author[aff1]{Petra Bosilj\corref{cor1}}
\author[aff1]{Başak Sakçak\corref{cor1}}

\address[aff1]{Department of Advanced Computing Sciences, Faculty of Science and Engineering, Maastricht University, Maastricht, The Netherlands}
\address[aff2]{Fine Field / Konijnenweg 12, 5962 BA Melderslo, The Netherlands}
\cortext[cor1]{\raggedright Correspondence: Nicolas Miroslav Husár (n.husar@student.maastrichtuniversity.nl)\;|\; Jelle Driessen (j.driessen@finefield.nl)\;|\; Petra Bosilj (petra.bosilj@maastrichtuniversity.nl)\;|\; Başak Sakçak (basak.sakcak@maastrichtuniversity.nl)}

\begin{abstract}
Mechanical blueberry harvesting requires a balance between removing ripe fruit and avoiding premature removal of unripe berries that could remain on the bush for a later harvest. However, operators generally lack real-time information about the maturity composition of the fruit being collected. This study develops an event-triggered machine-vision system that measures the maturity of harvested blueberries directly on the harvester. When a filled crate is moved by the machine, an RGB image is automatically captured and analysed using a three-class detector that identifies ripe, semi-ripe, and unripe berries. The system operates locally on the existing CPU-only industrial computer and therefore requires neither continuous image processing nor dedicated GPU hardware. During field testing, 79 crate images containing 48,634 visible blueberries in manually verified reference annotations were evaluated. The vision system estimated an overall ripe-fruit fraction of 90.22\%, compared with 90.97\% determined manually. Per-image ripe-fruit estimates showed a mean absolute error of 0.935 percentage points and a Pearson correlation of \(r=0.967\). Maturity estimates remained similar between the two sides of the harvester for which the light conditions varied despite poorer absolute counting performance under direct sunlight. All field events were processed within the available 14 s per-side crate interval. The proposed approach provides a practical means of quantifying one component of harvest selectivity while the machine is operating. Such information could support operator decision-making and, when combined with measurements of fruit recovery and losses, contribute to future feedback-based control of mechanical harvesting.
\end{abstract}


\end{frontmatter}

\section{Introduction}

Mechanical harvesting has become increasingly significant in blueberry production due to its ability to reduce reliance on seasonal human labour and facilitate rapid harvesting across large fields. The primary objective for growers extends beyond maximising the immediate weight of berries collected. It is to optimise the overall value of the field, which typically requires three harvests per season. For fresh-market production, machines must efficiently harvest fruit while preserving quality and minimising the collection of unripe berries that require further ripening. Previous studies show that machine design, harvesting intervals, and fruit handling affect fresh-market quality \citep{cai2021,devetter2019,sargent2020,sargent2021}.

The objective of this study is to develop and field-test an event-triggered computer-vision system that estimates the maturity composition of blueberries directly on a mechanical harvester while harvesting is in progress. A key challenge in current harvesting operations is the lack of immediate feedback on the maturity composition of collected fruit. Commercial sorter and packout results provide valuable information about harvested product quality, but these measurements are generally available only after harvesting. Similar real-time monitoring approaches have been investigated for other harvested crops \citep{jahari2015,jahari2017,chen2020,wu2024,pan2026}.

Computer vision has also been widely investigated in blueberry production, including maturity classification, yield estimation, and machine-mounted monitoring \citep{li2014,yang2014,esaki2025,maceachern2023,liu2023,zhao2025,das2023,haydar2023,mullins2026}. However, the real-time assessment of blueberry maturity after mechanical detachment remains comparatively unexplored.

The main contribution of this work is the integration of event-triggered crate imaging, three-class blueberry maturity detection, and local CPU inference into a field-deployed monitoring system. The approach is evaluated using 79 field-acquired crate images with manually annotated blueberries, examining maturity-estimation accuracy and processing performance under operating conditions. Rather than introducing a new detection architecture or implementing automatic machine control, the study investigates whether harvested-fruit maturity can be measured sufficiently accurately and promptly to support future feedback-based harvesting. Because maturity composition alone does not measure fruit remaining on the bush or losses to the ground, it is considered one component of harvest selectivity rather than a complete measure of harvesting efficiency.

\section{Related Work}

Previous research on computer vision in blueberry production has primarily focused on maturity assessment and yield estimation before harvesting. Early studies used RGB imaging to identify different blueberry growth stages \citep{li2014}, while hyperspectral imaging was investigated for distinguishing fruit maturity under field conditions \citep{yang2014}. More recent research has employed deep-learning approaches, particularly YOLO-based \citep{redmon2016yolo} detectors, for maturity classification, berry counting, and yield estimation \citep{esaki2025,maceachern2023}. Similar approaches have also been applied to ripeness detection for robotic picking \citep{liu2023} and three-dimensional fruit localisation \citep{zhao2025}. Although these studies demonstrate the feasibility of visual maturity assessment under outdoor conditions, they primarily examine berries still attached to the plant rather than fruit collected by mechanical harvesters.

Real-time monitoring of harvested-product quality has also been investigated in other agricultural applications. \citet{jahari2015,jahari2017} developed machine-vision systems to inspect harvested paddy for damaged grain and unwanted material, including the presentation of quality information to operators during harvesting. Similar systems have been developed to monitor grain breakage and impurity content in combine harvesters \citep{chen2020,wu2024}. In potato harvesting, computer vision has been applied to detect potatoes and soil clods, estimate tuber dimensions, and identify damaged products \citep{almallahi2010,blok2025,pan2026}. Together, these studies demonstrate that harvested-product characteristics can be monitored during machine operation and potentially used to support adjustments to harvesting settings.

Machine-mounted vision systems have also been developed specifically for blueberry harvesting. \citet{das2023} used RGB cameras to detect berries and unwanted material, including leaves, stems, and dirt, on a commercial wild-blueberry harvester. \citet{haydar2023} employed an OAK-D camera and deep-learning methods to estimate fruit height and automatically adjust harvester-header positioning. More closely related to the present study, \citet{mullins2026} developed a real-time yield-monitoring system using a three-dimensional time-of-flight camera to estimate the mass of harvested blueberries in totes while accounting for debris. However, their objectives focus on debris detection, machine positioning, and harvested mass rather than the maturity composition of the collected fruit.

Beyond maturity assessment, previous research has also examined the effects of mechanical harvesting on blueberry quality. Studies of modified over-the-row harvesters have shown that catching surfaces, harvesting configurations, and fruit handling influence firmness, bruising, and fresh-market suitability \citep{devetter2019,sargent2020,sargent2021}. Postharvest sensing methods, including hyperspectral imaging, have also been investigated for predicting fruit firmness and identifying internal bruising \citep{leiva2013,fan2018}. These properties are important for determining commercial quality but differ from colour-based maturity composition.

\section{Materials and Methods}

\subsection{Study context and harvester configuration}

\begin{figure}%[t]
    \centering
    \includegraphics[width=\columnwidth]{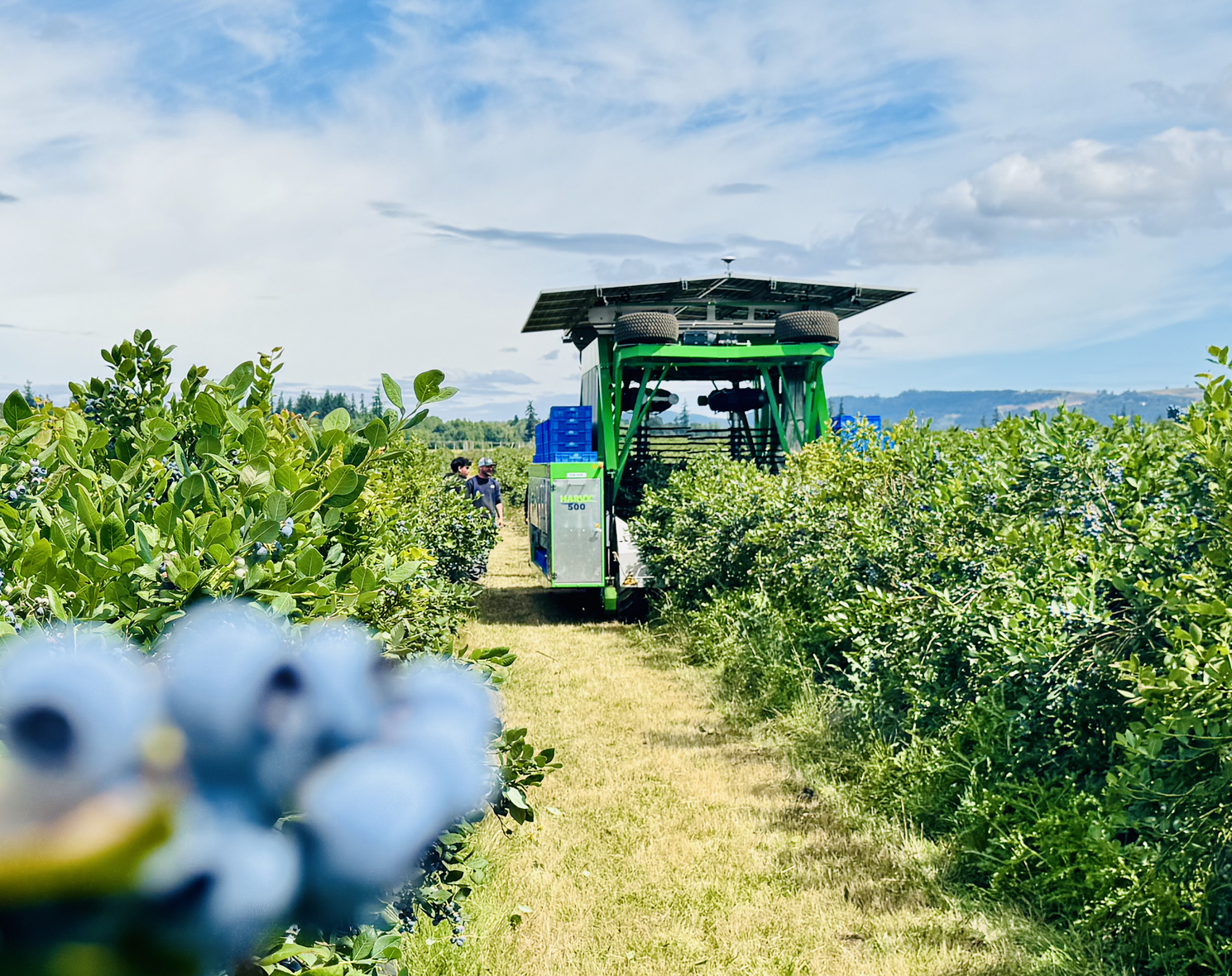}
    \caption{FineField Harvy 500 mechanical blueberry harvester operating in a blueberry field. Photograph taken on a different day from the reported field experiment.}
    \label{fig:harvester}
\end{figure}

The system was tested on a FineField Harvy 500 mechanical blueberry harvester (Fig.~\ref{fig:harvester}) in Weeze, Germany, on 21 July 2026. The harvester operates bidirectionally, meaning that its direction of travel can be reversed. Therefore, the two sides are not designated as left and right, but instead as the blue side and the white side. Each side operates independently, with its own crate-filling system, crates, crate pusher, and fixed camera positioned above the filled crate. The cameras do not observe the same crate from different angles; instead, each camera records a separate crate from its respective side of the machine.

During harvesting, blueberries are removed from the bush using shaker heads and transported to the crate using a catching surface that directs the berries into small buckets, which then drop them into the crate. When a crate reaches a set weight threshold (in this case 6.5 kg), it is pushed and replaced by an empty crate.

\subsection{Camera position and field of view}

\begin{figure}[t]
    \centering
    \includegraphics[width=\columnwidth]{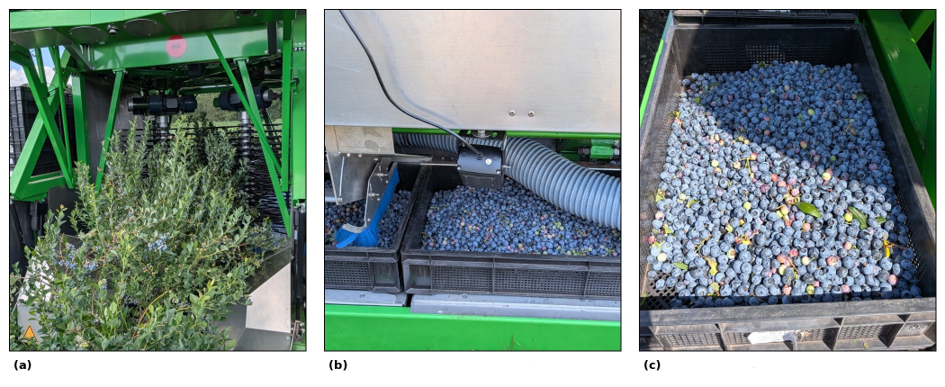}
    \caption{Field context of the study: (a) blueberry crop entering the mechanical picking zone; (b) machine-mounted camera positioned above the crate stream; and (c) harvested product showing ripe, semi-ripe, unripe fruit and debris.}
    \label{fig:field_context}
\end{figure}

The same camera settings and angle were used on both the blue and white sides. The cameras were positioned approximately 20~cm above the crate, meaning we used wide-field-of-view lenses. The crop, camera installation, and collected fruit are illustrated in Fig.~\ref{fig:field_context}.

The camera system used a 5 MP, 1/1.8-inch CMOS global-shutter sensor. A global shutter was selected because it captures the entire image simultaneously, helping reduce image distortion caused by movement or vibration. The camera module can provide images at up to 2448 $\times$ 2048 pixels at 30 frames per second and supports manual or automatic white balance, shutter speeds from 1/12 s to 1/100,000 s, backlight compensation, strong-light suppression, noise reduction, and digital wide dynamic range. Its specified operating temperature is $-$20 \textdegree{}C to +60 \textdegree{}C.

For installation on the harvester, the cameras were placed in weather-resistant housings rated IP67 or higher and connected through active Power over Ethernet (PoE). This allowed both power and network communication to be provided via the camera connection, reducing the amount of wiring required on the machine.

Only the top layer of the berries is visible in each image. There was no space for the camera above the crate that is currently being filled, as the leaf blower would blow dirt and leaves into the camera. Instead, each image is treated as a sample of the harvested product.

\subsection{Event-triggered image capture}

Image capture was synchronised with the crate-handling system rather than performed continuously. When a filled crate was pushed from the collection position, a sensor triggered a signal sent to the industrial PC. The trigger also identified which side of the harvester had pushed the crate. After receiving the signal, the software waited one second before capturing an image from the corresponding camera. This delay allowed the crate to move into a stable position before the image was taken.

This event-triggered approach reduced unnecessary image processing and created a direct link between the machine event and the recorded sample. Each trigger produced one timestamped crate image and one corresponding maturity-analysis result. The system also recorded capture events in log files, enabling the identification of missed or incorrectly timed images.

\subsection{Datasets and Annotations}

The model was developed using two publicly available blueberry datasets and photographs provided by Fine Field. The Blueberry Detection Dataset (BDD) \citep{liBlueberryDetectionDataset} contains images of blueberries with bounding-box annotations. Its OPENset subset was used to train the initial single-class detector. This training configuration contained 79 public images and one additional crate photograph, for a total of 80 images with 4,248 annotated blueberries. The images were divided into 59 training, 16 validation, and 5 test images. They had different original resolutions, while the detector used a $640\times640$-pixel input during training.

\begin{figure}[t]
    \includegraphics[width=\columnwidth]{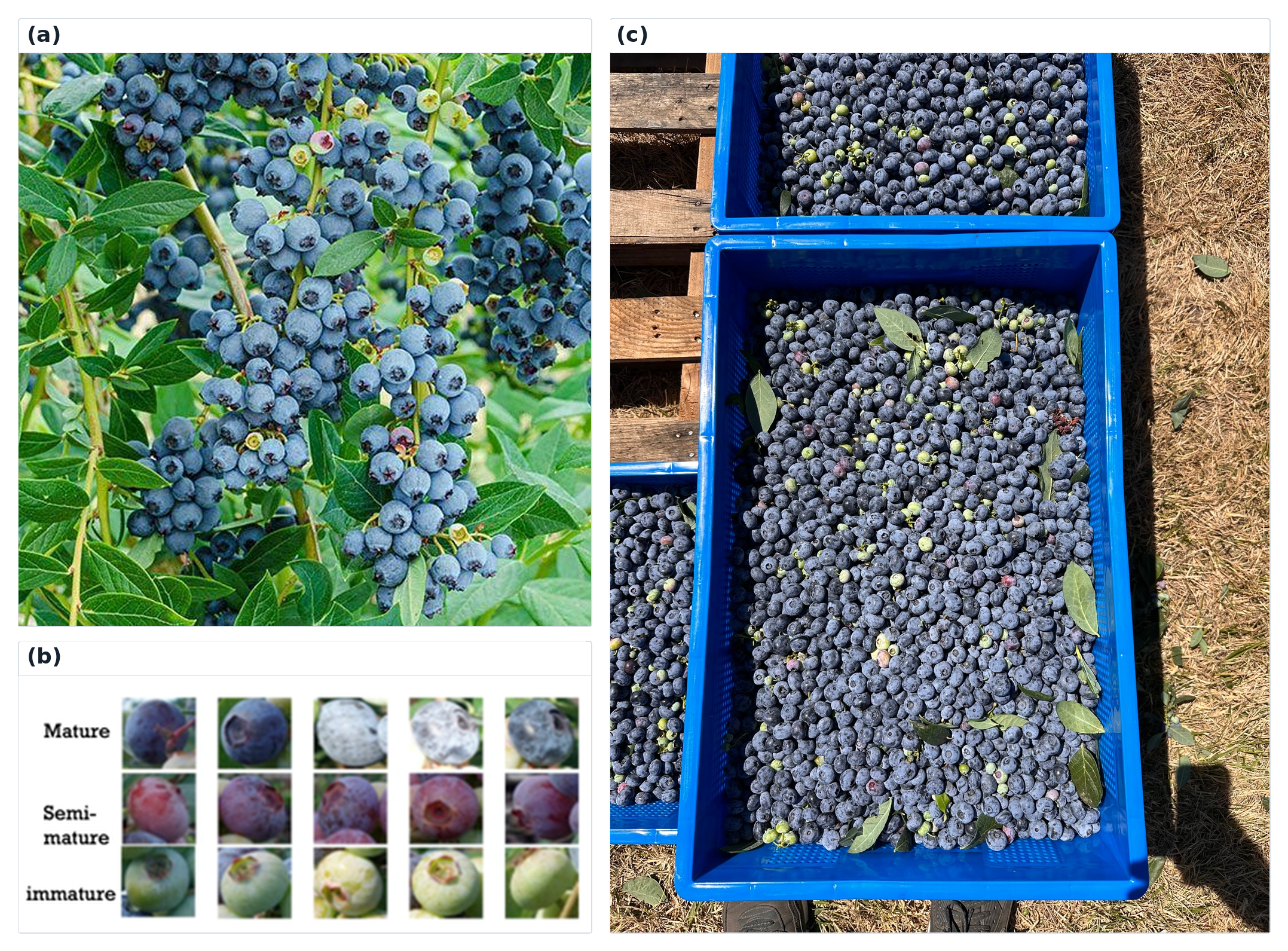}
    \caption{Example images from the three datasets: (a) clustered blueberries from the Blueberry Detection Dataset (BDD) \citep{liBlueberryDetectionDataset}; (b) individual berries at different maturity stages from the Blueberry Maturity Classification Dataset (BMCD) \citep{liBlueberryMaturityDataset}; and (c) harvested blueberries collected in Fine Field crates.}
    \label{fig:dataset_examples}
\end{figure}

The second source was the Blueberry Maturity Classification Dataset (BMCD) \citep{liBlueberryMaturityDataset}, which contains images of individual berries representing different maturity stages. These images were used to train a preliminary YOLO26s \citep{jocher2026yolo26} model to distinguish ripe, semi-ripe, and unripe berries. The preliminary model was then applied to a selected collection of 106 images, consisting of 96 images from the BDD dataset and 10 additional photographs provided by Fine Field. The BDD images were selected because they contained many closely spaced and overlapping blueberries on bushes, resembling the densely packed berries in the harvester crates. The existing bounding-box annotations were retained, and the preliminary model was used to assign maturity classes to the berries. The predicted classes were then reviewed and corrected in the Computer Vision Annotation Tool (CVAT) \citep{cvat_software}. The Fine Field images were manually annotated where bounding boxes were not already available. The resulting three-class annotations were used to train the final maturity-detection model. The recorded final training configuration contained 8,455 bounding-box annotations, including 7,587 ripe, 571 unripe, and 297 semi-ripe berries. Ripe berries were mainly blue, unripe berries mainly green, and semi-ripe berries showed intermediate colouration. The dataset configurations are summarised in Table~\ref{tab:datasets}. Example images from the three data sources are shown in Fig.~\ref{fig:dataset_examples}.

The field evaluation used 79 filled-crate images: 40 from the blue side and 39 from the white side. The cameras captured images at 1920 × 1080 pixels. These images were not used during model training. The sun-affected images were retained in the dataset because they still contained useful information and reflected the real conditions encountered during harvesting. Only berries whose maturity could be visually determined were included in the reference annotations. Berries in severely overexposed regions, where their maturity could not be reliably identified, were excluded. The machine normally produces a filled crate approximately every 14 seconds per side, while the blue and white sides operate independently. In a situation where both sides require processing within the same period, two images would need to be analysed within approximately 14 seconds. Commercial Class I packout results for two corresponding lots were considered as a downstream comparison after approximately six weeks of controlled-atmosphere cold storage. Individual imaged crates were not tracked through storage and sorting.

\begin{table*}[t]
\centering
\caption{Dataset configurations used during model development and field evaluation. Counts for expanded training datasets include augmented files, whereas the original-image count excludes augmented copies.}
\label{tab:datasets}
\small
\begin{tabularx}{\textwidth}{@{}p{3.6cm}rrrX@{}}
\toprule
Configuration & Image files & Originals & Boxes & Notes \\
\midrule
Initial single-class detection & 80 & 80 & 4,248 & 59 training, 16 validation, 5 test; one blueberry class \\
Final three-class maturity detection & 106 & 106 & 8,455 & 96 selected BDD images and 10 Fine Field photographs; three maturity classes \\
Field evaluation & 79 & 79 & 48,634 & 40 blue-side and 39 white-side images; $1920\times1080$ pixels \\
\bottomrule
\end{tabularx}
\end{table*}

\subsection{Model Training}

The final detector was developed in two main phases using YOLO26s \citep{jocher2026yolo26}. The initial model used weights pretrained on the COCO dataset \citep{lin2014coco}. In the first phase, the model was trained to detect blueberries as a single class, independently of maturity. In the second phase, the resulting detector was adapted to distinguish three maturity classes: ripe, semi-ripe, and unripe. The final model therefore locates individual blueberries and assigns their maturity classes in a single inference step.

Training of the single-class detector was configured for up to 100 epochs, with a batch size of 16, an input resolution of $640\times640$ pixels, and early stopping with a patience of 20 epochs. For the maturity-detection, the single-class model checkpoint was used to initialise the three-class detector. The final model was trained using Ultralytics 8.4.16, with 100 epochs, a batch size of 6, and an input resolution of $960\times960$ pixels. The input resolutions refer to the images supplied to the model rather than the original resolution of every photograph. Images were resized to the configured input dimensions, with padding applied where required to preserve their aspect ratio. Consequently, smaller exported images could be enlarged, while larger source images were reduced to fit the model input. Image preprocessing followed the Ultralytics procedure \citep{ultralyticsPreprocessingDocs}.

The optimiser was selected automatically by Ultralytics \citep{ultralyticsTrainDocs}. AdamW \citep{loshchilov2019decoupled} was used with effective initial learning rates of approximately 0.002 for the initial single-class detector and 0.001429 for the final three-class detector. The learning rate of 0.001 specified in the final training script was overridden by automatic optimiser selection. The final run used an AdamW first-moment coefficient of 0.9 and an effective weight decay of approximately $5.16\times10^{-4}$. The learning rate was not selected through a separate comparison of candidate values. A fixed random seed of 42, deterministic execution, and automatic mixed precision were enabled. The training script requested two warm-up epochs.

Data augmentation was used to introduce variation in berry appearance, size, and position. Colour augmentation was reduced during three-class fine-tuning because visible colour was an important cue for distinguishing maturity stages. The hue, saturation, and value (HSV) transformations followed the Ultralytics implementation \citep{ultralyticsAugmentDocs}. Hue was shifted by a randomly sampled fraction of the colour wheel, while saturation and brightness were adjusted using randomly sampled multiplicative factors, with values clipped to the valid image range. For the final model, the hue shift was limited to $\pm0.010$ of the colour wheel, equivalent to approximately $\pm3.6^\circ$. Saturation was multiplied by a factor between 0.50 and 1.50, and brightness by a factor between 0.70 and 1.30. These reduced settings were intended to introduce variation in illumination and colour appearance while limiting changes that could obscure the distinction between ripe, semi-ripe, and unripe berries. The complete augmentation settings for the initial single-class and final three-class training stages are summarised in Table~\ref{tab:augmentations}.

\begin{table*}[t]
\centering
\caption{Augmentation settings for the initial single-class detector and final three-class maturity detector. Mosaic and flipping values are application probabilities. Scaling, translation, and HSV settings specify the magnitude of randomly sampled transformations. Initial settings were Ultralytics defaults.}
\label{tab:augmentations}
\small
\begin{tabularx}{\textwidth}{@{}p{3.05cm}p{2.15cm}p{2.15cm}X@{}}
\toprule
Augmentation & Single-class model & Three-class model & Interpretation \\
\midrule
Mosaic & 1.0 & 0.5 &
Applied with 100\% and 50\% probability, respectively, while enabled \\

Close mosaic & Last 10 epochs & Last 10 epochs &
Mosaic disabled during the final 10 epochs \\

Scale & 0.5 & 0.3 &
Random scale factors of 0.50--1.50 and 0.70--1.30, respectively \\

Translation & 0.10 & 0.05 &
Random horizontal and vertical shifts of up to $\pm10\%$ and $\pm5\%$ of the image dimensions \\

HSV hue & 0.015 & 0.010 &
Random hue shifts of up to $\pm1.5\%$ and $\pm1.0\%$ of the colour wheel ($\pm5.4^\circ$ and $\pm3.6^\circ$) \\

HSV saturation & 0.70 & 0.50 &
Random saturation multipliers of 0.30--1.70 and 0.50--1.50, respectively, with clipping \\

HSV brightness (value) & 0.40 & 0.30 &
Random brightness multipliers of 0.60--1.40 and 0.70--1.30, respectively, with clipping \\

Horizontal flip & 0.50 & 0.50 &
Applied with 50\% probability \\

Vertical flip & 0 & 0 &
Disabled \\

Rotation, shear, perspective & 0 & 0 &
Disabled \\

MixUp and copy-paste & 0 & 0 &
Disabled \\
\bottomrule
\end{tabularx}
\end{table*}

\subsection{Inference pipeline and maturity-composition estimate}

For each captured crate image, the software prepares the image, runs the YOLO detector, and stores the resulting bounding boxes, maturity classes, and confidence values. The detections are then counted by maturity class to calculate the maturity-composition estimate.

For image (i), let:

$(M_i)$ = number of ripe berries,

$(S_i)$ = number of semi-ripe berries,

$(I_i)$ = number of unripe berries.

The total number of detected berries is:

\begin{equation}
T_i = M_i + S_i + I_i
\label{eq:total_detections}
\end{equation}

The detected ripe fruit fraction is:

\begin{equation}
P_i = 100\,\frac{M_i}{T_i}
\label{eq:ripe_fraction}
\end{equation}

This value describes the maturity composition of berries that are visible in the image and detected by the model.

\subsection{Data logging and traceability}

For each captured crate, the system stores the image together with the corresponding analysis results. The saved information includes the camera side, capture identifier, timestamp, image size, ripe count, semi-ripe count, unripe count, total number of detected berries, and the calculated maturity-composition estimate. The system also stores individual detections, including their class, confidence score, and bounding-box coordinates. Runtime performance was evaluated from the logs recorded during the 79-event field experiment.

\subsection{Deployment and OpenVINO inference}

The final detection system was run on the industrial computer installed on the harvester rather than on the more powerful computer used during model development. The industrial PC contains an Intel Celeron J1900 processor running at approximately 1.99 GHz, with four CPU cores and approximately 3.7 GB of RAM. No GPU was used during the reported deployment.

The trained YOLO26s model was exported to OpenVINO (Open Visual Inference and Neural network Optimization), a toolkit for executing deep-learning models efficiently on hardware including Intel CPUs \citep{openvinoToolkitDocs}. The deployed model used a $1280\times1280$-pixel input, although the final model had been trained using $960\times960$-pixel inputs. The larger input was selected using a preliminary comparison on one Fine Field crate image held out from the model-development training, validation, and test splits. The 1280-pixel input produced the lowest error on that image.

\subsection{Overview of Experiments}
The evaluation was divided into three main parts: model development, maturity-estimation performance, and runtime performance. First, the training and validation results of the initial single-class and final three-class detectors were examined. The effect of model input resolution was also compared using a single Fine Field crate image that was not included in training, validation, or testing, with input sizes of $640\times640$, $960\times960$, and $1280\times1280$ pixels.

The field evaluation was conducted using 79 crate images collected from both sides of the harvester. Detection performance, absolute berry counts, and maturity-composition estimates were compared with the reference annotations, including an assessment of the effect of sunlight. Runtime performance was evaluated using a representative-image benchmark and processing logs recorded during field operation. In addition, a heuristic estimate of the total number of berries per crate and the commercial packout results were included as supplementary comparisons.

\subsection{Evaluation Metrics}

Object-detection performance was evaluated using mean average precision (mAP), precision, and recall. The mAP@50 metric measures average precision at an intersection-over-union (IoU) threshold of 0.50, while mAP@50--95 averages the results over thresholds from 0.50 to 0.95. Precision is the proportion of predicted detections that are correct, and recall is the proportion of reference objects detected. The F1 score is the harmonic mean of precision and recall, and macro-F1 is the average of the three class-wise F1 scores.

The ripe-fruit fraction was compared with the reference annotations using mean absolute error (MAE), root mean squared error (RMSE), mean bias, and Pearson's correlation coefficient ($r$). MAE is the average absolute difference between the predicted and reference fractions, RMSE gives greater weight to larger errors, and mean bias is the average signed difference. Pearson's $r$ measures the linear relationship between the two fractions. Ripe-fraction errors are given in percentage points. Absolute counts were compared using count error and mean absolute percentage error (MAPE), which expresses the average absolute counting difference relative to the reference count. For runtime results, P95 denotes the 95th percentile of the recorded times. Differences in per-image absolute ripe-fraction error between the two harvester sides were tested using Welch's two-sample test, which does not assume equal variances.

\section{Results and Discussion}

\subsection{Model-Development and Input-Resolution Results}

The first single-class blueberry detector achieved a validation mAP@50 of 0.951 and mAP@50--95 of 0.688 on the 16 validation images. These images were separate from the training images but were used to select the best checkpoint. The values therefore describe model-development validation rather than independent final testing.

The final three-class training run reached an mAP@50 of 0.772, mAP@50--95 of 0.704, precision of 0.932, and recall of 0.721. As this model was intended for deployment, all available annotated training images were used for fine-tuning rather than reserving a separate validation subset. The same images were therefore also used for evaluation. Consequently, these metrics describe training-set performance rather than independent validation results. The final model was evaluated separately on the field-acquired crate images.

The effect of input resolution on maturity estimation and inference time is summarised in Table~\ref{tab:resolution_results}. Higher input resolutions reduced the ripe-fraction error in the single-image comparison but increased processing time in the separate timing benchmark. The $1280\times1280$-pixel input produced the lowest error and was therefore selected for deployment. However, the accuracy comparison was performed using only one Fine Field crate image, so the results do not establish whether the same improvement would be observed across other images.

\begin{table}[t]
\centering
\caption{Absolute ripe-fraction error from the single-image comparison and inference times from the separate resolution benchmark.}
\label{tab:resolution_results}
\small
\begin{tabular}{@{}lcc@{}}
\toprule
Input size (pixels) & Error (pp) & Inference (ms) \\
\midrule
$640\times640$ & 8.01 & 2170 \\
$960\times960$ & 2.60 & 2645 \\
$1280\times1280$ & 2.24 & 4000 \\
\bottomrule
\end{tabular}
\end{table}

A higher input resolution may help preserve information about small and partly hidden berries, although this comparison did not examine which individual detections changed with resolution. The chosen input size therefore reflects the balance between the observed maturity estimate and the processing time available after each crate trigger, rather than a resolution shown to be optimal under all conditions.

\subsection{Field Maturity and Detection Performance}

\begin{figure}[t]
    \centering
    \includegraphics[width=0.99\linewidth]{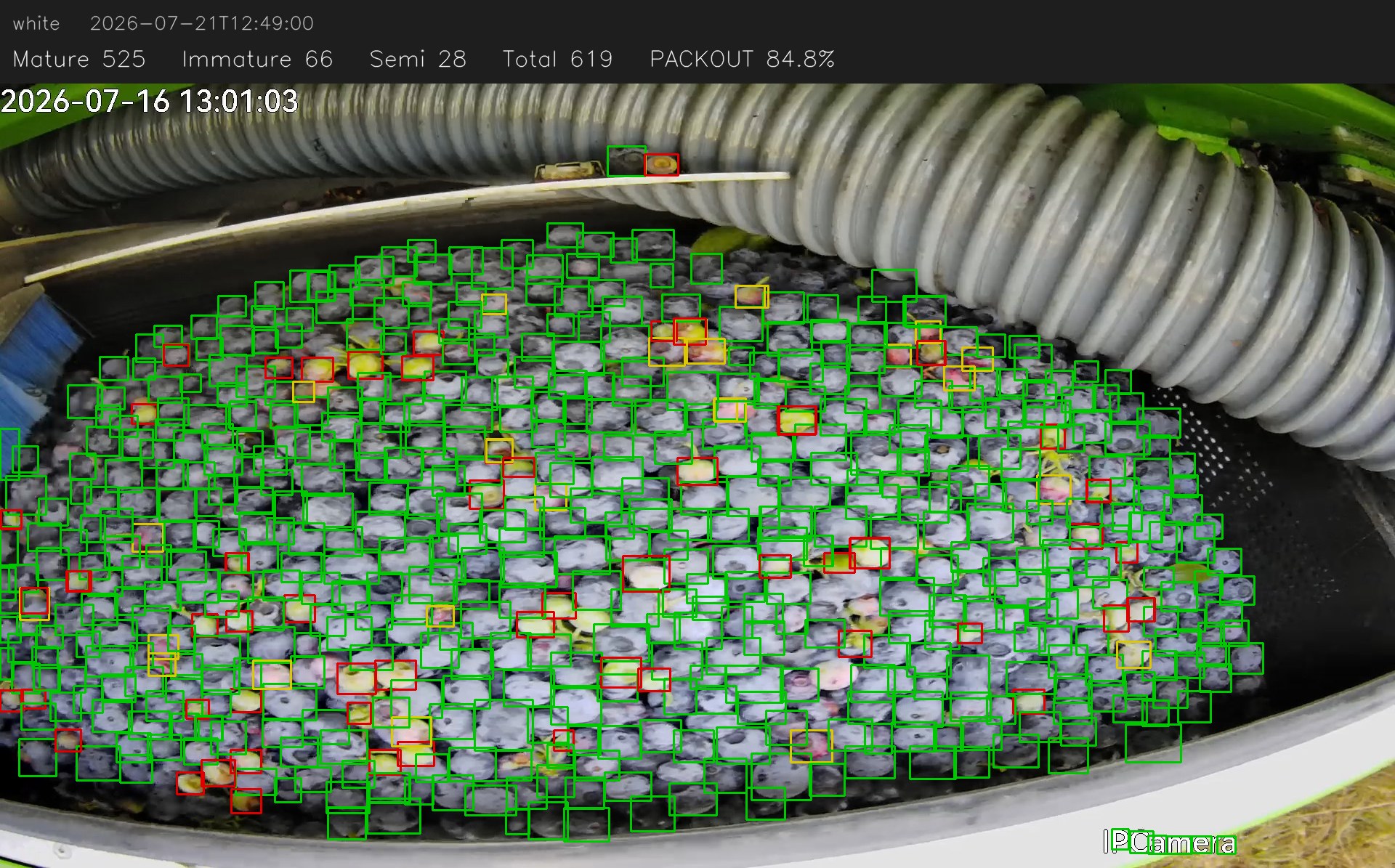}
    \caption{Example of the system's saved detection output for a white-side crate. Green, yellow, and red bounding boxes indicate ripe, semi-ripe, and unripe blueberries, respectively. The displayed ``PACKOUT'' value of 84.8\% is the detected ripe-fruit fraction (525 of 619 detections), not commercial Class~I packout.}
    \label{fig:crate_detection_example}
\end{figure}

An example of the saved detection output from the white-side camera is shown in Fig.~\ref{fig:crate_detection_example}. For this crate, the detector identified 525 ripe, 28 semi-ripe, and 66 unripe berries, giving 619 detections and a ripe-fruit fraction of 84.8\%. This is one crate result rather than the overall maturity estimate.

\begin{figure}[t]
    \centering
    \includegraphics[width=\columnwidth]{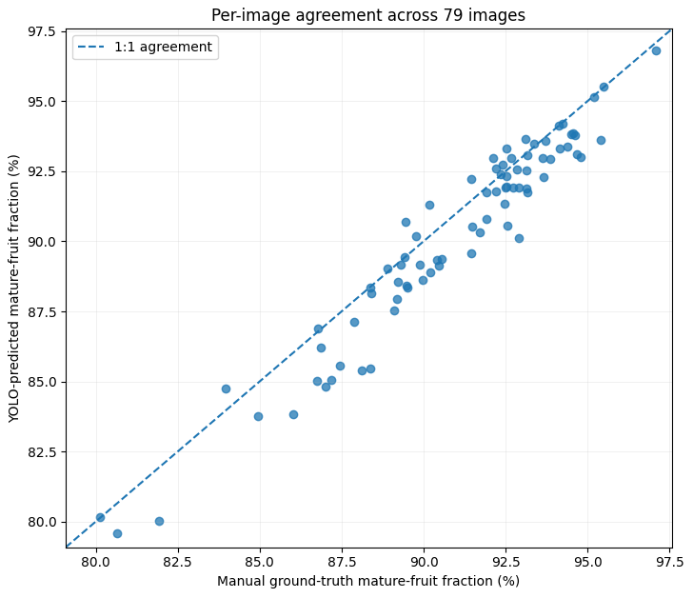}
    \caption{Predicted versus reference ripe-fruit fraction for 79 field-acquired crate images. The dashed line represents perfect 1:1 agreement.}
    \label{fig:maturity_validation}
\end{figure}

Across the 79 analysed field images, the detector estimated an overall ripe-fruit fraction of 90.22\%, compared with 90.97\% from the reference annotations, a difference of $-0.75$ percentage points. The per-image mean absolute error was 0.935 percentage points and Pearson's correlation was $r=0.967$ ($R^2=0.935$). The mean bias was $-0.719$ percentage points and the maximum absolute per-image error was 2.84 percentage points. Of the 79 images, 45 (57.0\%) were within $\pm1$ percentage point, 64 (81.0\%) were within $\pm1.5$ percentage points, and 73 (92.4\%) were within $\pm2$ percentage points. The overall and side-specific detection results are given in Table~\ref{tab:perception_side_comparison}, and the per-image ripe-fraction agreement is shown in Fig.~\ref{fig:maturity_validation}.

\begin{table}[t]
\centering
\caption{Detection and maturity-estimation performance across the field images and harvester sides. Errors in ripe-fruit fraction are reported in percentage points (pp).}
\label{tab:perception_side_comparison}
\small
\resizebox{\columnwidth}{!}{%
\begin{tabular}{lccc}
\toprule
\textbf{Metric} & \textbf{Overall} & \textbf{White side} & \textbf{Blue side} \\
\midrule
Number of images                  & 79      & 39      & 40      \\
Reference berries              & 48,634  & 25,575  & 23,059  \\
\midrule
Macro-F1                          & 0.862   & 0.864   & 0.859   \\
F1 ripe                         & 0.950   & 0.960   & 0.940   \\
F1 semi-ripe                    & 0.803   & 0.792   & 0.815   \\
F1 unripe                       & 0.833   & 0.842   & 0.823   \\
\midrule
Count error (\%)                  & +6.00   & +5.15   & +6.94   \\
Reference ripe fraction (\%)       & 90.97   & 90.95   & 91.00   \\
Predicted ripe fraction (\%)    & 90.22   & 90.16   & 90.29   \\
Ripe-fraction MAE (pp)          & 0.935   & 0.998   & 0.872   \\
Ripe-fraction RMSE (pp)         & 1.165   & 1.241   & 1.086   \\
Mean bias (pp)                    & -0.719  & -0.771  & -0.667  \\
Pearson $r$                       & 0.967   & 0.971   & 0.964   \\
\bottomrule
\end{tabular}%
}
\end{table}

The detector produced 51,552 berry detections compared with 48,634 reference annotations, an overall count difference of +6.00\%. For ripe berries, the predicted and reference counts were 46,510 and 44,243 (+5.12\%); for unripe berries they were 3,624 and 3,101 (+16.87\%); and for semi-ripe berries they were 1,418 and 1,290 (+9.92\%). The larger relative overcount for the less frequent classes is consistent with the negative bias in ripe-fruit fraction, even though the fraction was less affected than the absolute number of detections.

The blue-side camera experienced direct sunlight and partial overexposure, while the white side had more uniform lighting. The mean absolute percentage error of the berry counts was 7.80\% on the blue side and 5.64\% on the white side, although reported recall remained above 97\% for both streams. In contrast, ripe-fraction MAE was 0.87 percentage points on the blue side and 1.00 percentage points on the white side, with no statistically significant difference in per-image absolute error ($p=0.39$, Welch's two-sample test). These are side-specific observations rather than a controlled isolation of sunlight, and the reference annotations excluded berries whose maturity could not be determined in severely overexposed regions.

The system is mainly intended to estimate the maturity of harvested blueberries, rather than count every berry accurately. Although the detector counted more berries than were present in the reference annotations, the estimated ripe-fruit fraction remained close to the manually determined value. This shows that errors in berry counting do not necessarily lead to large errors in maturity estimation. However, the larger counting errors for unripe and semi-ripe berries indicate that detecting these classes remains important.

Direct sunlight made it more difficult to detect individual berries on the blue side of the harvester. Adding a shade around the crate or using controlled lighting could improve image quality and make detection more consistent, as also suggested in previous agricultural monitoring studies \citep{chen2020,jahari2015,jahari2017}. The system currently measures the maturity of visible harvested berries, but it does not measure how many berries remain on the bush or how much fruit is lost during harvesting.

\subsection{Heuristic Estimate of Berries per Crate}

The approximate number of blueberries in each 6.5~kg crate was estimated using two complementary approaches: one based on berry size, density, and crate mass, and another based on the estimated volume of the berry pile.

For the first approach, the median berry diameter was estimated from the widths of the YOLO bounding boxes, converted from pixels to centimetres using the known crate dimensions. The resulting median diameter was approximately 1.67~cm. Assuming that blueberries are approximately spherical, the volume of an individual berry was calculated as:

\begin{equation}
V_b = \frac{\pi}{6}d^3
\label{eq:berry_volume}
\end{equation}

where $d$ is the estimated berry diameter. This resulted in an individual berry volume of approximately 2.44~$\mathrm{cm^3}$. Using a density of 1023~$\mathrm{kg\,m^{-3}}$, as reported for fresh commercial blueberries by \citet{nuangjamnong2026}, the estimated mass of one berry was approximately 2.49~g. Dividing the total crate mass of 6.5~kg by this value yielded an estimate of approximately 2600 berries per crate.

The second approach estimated the number of berries from the volume of the berry pile. The crate dimensions were approximately $60 \times 40$~cm, and the pile was modelled as a base layer with a thickness equal to one berry diameter, topped by an elliptical paraboloid. The maximum pile height was estimated at approximately 9~cm, gradually decreasing toward the edges. The total pile volume was calculated as:

\begin{equation}
V_{\mathrm{pile}} = LWd + \frac{\pi ab(h-d)}{2}
\label{eq:pile_volume}
\end{equation}

where $L$ and $W$ are the crate length and width, $a=L/2$ and $b=W/2$ are the semi-axes of the elliptical paraboloid, $d$ is the estimated berry diameter, and $h$ is the maximum pile height.

This approximation resulted in a total pile volume of approximately 10.9~L. Assuming a bulk porosity of 0.40, based on \citet{nuangjamnong2026}, approximately 60\% of this volume was occupied by berries, corresponding to 6.54~L. Dividing this volume by the estimated individual berry volume of 2.44~$\mathrm{cm^3}$ yielded approximately 2680 berries per crate.

The two approaches produced similar estimates of approximately 2600 and 2680 berries per crate. Based on these results, a representative value of approximately 2650 berries per 6.5~kg crate was adopted for interpreting the image-based results. This value should be considered a heuristic estimate rather than a direct count, as berry size, shape, packing density, and pile geometry may vary between crates.

The reference annotations contained an average of approximately 616 visible berries per crate, corresponding to about 23\% of the heuristic total. The camera therefore measures only the upper surface of the harvested fruit. Agreement between the two estimates provides a useful order-of-magnitude check, but does not establish that the visible berries represent a random sample of the full crate or that the maturity composition below the surface is the same.

\subsection{CPU Runtime and Event Handling}

For one representative image processed on the industrial PC, preprocessing required approximately 70.6~ms, inference 3882.1~ms, and berry counting 1.2~ms, giving a total processing time of approximately 3953.9~ms. This single-image benchmark is separate from the field-event timing results.

For the 79 analysed field events, mean model inference time was 4.32~s and mean recorded processing time, including acquisition and post-processing, was 4.93~s. Trigger-to-result latency had a median of 5.84~s and a maximum of 11.13~s (Table~\ref{tab:runtime_field}). All 79 analysed events produced a result within the available 14~s per-side crate interval.

\begin{table}[t]
\centering
\caption{Runtime performance measured from the 79 trigger events. Trigger-to-result latency includes the intentional 1~s settling delay and any waiting caused by overlapping events.}
\label{tab:runtime_field}
\begin{tabular}{lrrrr}
\hline
\textbf{Metric} & \textbf{Mean} & \textbf{Median} &
\textbf{P95} & \textbf{Max} \\
\hline
Capture (s)            & 0.53 & 0.36 & 1.12 & 1.22 \\
Inference (s)          & 4.32 & 4.31 & 4.47 & 4.50 \\
Processing (s)         & 4.93 & 4.86 & 5.48 & 5.59 \\
Trigger-to-result (s)  & 6.24 & 5.84 & 9.40 & 11.13 \\
\hline
\end{tabular}
\end{table}

Because the two crate streams operated independently, trigger events sometimes overlapped. Of the 78 intervals between recorded events, 28 (35.9\%) were shorter than 14~s and 14 (17.9\%) were shorter than 5~s. The closest blue/white trigger pair was separated by 0.101~s; its two results became available after 6.44~s and 11.13~s. These observations show that the deployed pipeline handled overlapping requests during the analysed session, but do not establish long-term throughput under other operating conditions.

In this application, trigger-to-result time is more relevant than continuous-video frame rate because an image is required only when a filled crate is moved. The field measurements show that the existing industrial computer was able to serve both independent crate streams during the observed test period without dedicated GPU hardware. Longer trials would be needed to assess the available processing margin when crate flow increases or other software tasks run on the same computer.

\subsection{Downstream Commercial Packout}

Commercial Class~I packout results were available for two lots of blueberries harvested on the same day as the field experiment. Class~I packout refers to the proportion of harvested fruit that meets commercial quality requirements for sale after sorting. Unlike the ripe-fruit fraction estimated by the vision system, this grading also considers properties such as firmness, bruising, size, appearance, decay, and foreign material. Together, the two lots contained approximately 5144~kg of fruit and achieved Class~I packout rates of 83.3\% and 85.6\%. Weighted by lot mass, the combined Class~I packout was approximately 83.8\% after six weeks of controlled-atmosphere storage.

The commercial packout was lower than the ripe-fruit fraction measured during harvesting. However, these values describe different characteristics of the fruit. The vision system estimates the proportion of visibly ripe berries based mainly on colour, while commercial packout measures the proportion that remains suitable for Class~I sale after storage and sorting. Properties such as internal bruising and firmness cannot be reliably determined from the RGB images used in this study \citep{fan2018,leiva2013}. Previous research also shows that commercial quality can change during storage. For example, \citet{forney2022} reported an average of 88.5\% marketable fruit after storing five highbush blueberry cultivars for six weeks at 1~$^\circ$C under controlled-atmosphere conditions. Although their marketability criteria differ from the Class~I grading used here, the study provides context for the effect of storage on fruit quality.

The 79 imaged crates represented roughly 513.5~kg of harvested fruit, while the commercial packout results covered a much larger quantity harvested on the same day. The individual crates were not tracked through storage and sorting, so the two measurements cannot be compared directly at crate level. The commercial results are therefore included to provide context for the harvest rather than to validate the vision system. The maturity measurement provides information about the composition of fruit collected by the harvester, but it should not be interpreted as a prediction of final commercial packout.

\section{Limitations}

The field evaluation was limited to 79 analysed images collected during one harvesting session on one machine. The results therefore do not establish performance under different cultivars, harvesting conditions, or seasons. The input-resolution comparison was also based on a single crate image, so it cannot determine the best resolution for all operating conditions. The training images were predominantly acquired on blueberry plants, while deployment images showed fruit densely packed in crates, and the maturity annotations were strongly imbalanced toward ripe berries.

The field reference annotations were prepared using initial detections from the evaluated model and then corrected in CVAT. Although missing berries and incorrect labels were reviewed, any remaining errors in these model-assisted annotations could make the reported detection and maturity-estimation agreement overly optimistic. Independent annotation of a representative subset of field images would provide a stronger assessment of detection accuracy.

Only berries visible at the top of each crate were evaluated. The surface layer may differ systematically from the berries underneath, and the heuristic estimate of the total berry count depends on assumptions about berry size, shape, density, packing and pile geometry. Consequently, the reported ripe-fruit errors describe agreement for visible annotated fruit, not the accuracy of the maturity fraction for the full crate. Severely overexposed berries whose maturity could not be determined were excluded from the reference annotations, which further limits conclusions about performance under direct sunlight.

The runtime results describe the observed workload on the tested CPU-only computer rather than guaranteed throughput at all harvesting rates. Finally, colour-based maturity composition is only one part of harvest quality. It does not establish commercial packout, fruit recovery, bruising, or the effect of changing shaker settings. These questions would require additional measurements and a longer field evaluation.

\section{Conclusions}

This study demonstrated event-triggered crate imaging and three-class blueberry maturity detection on a mechanical harvester using its existing CPU-only industrial computer. The field evaluation showed that maturity fractions derived from the detector were close to those calculated from the available reference annotations, and the recorded processing times were compatible with the observed crate flow. The system provides a measurement of visible harvested-fruit maturity, not a direct measure of full-crate maturity or commercial packout.

Future work should use independently verified field annotations, test more harvesting days and cultivars, and determine whether the top layer represents the whole crate. Better-controlled illumination and more maturity-labelled crate images could improve performance under difficult conditions. For automatic adjustment of the harvester, maturity measurements would also need to be combined with information about ripe fruit remaining on the bush and fruit losses. Existing approaches to debris detection, header control, and yield mapping provide related measurements that could support such a system \citep{das2023,haydar2023,maceachern2023,mullins2026}.

\section*{Acknowledgments}

This research was funded by Fine Field.

\section*{Data Availability Statement}

The data that support the findings of this study
are available from the corresponding author
upon reasonable request.

\section*{Declaration of Generative AI and AI-Assisted Technologies}

During the preparation of this manuscript, the authors used OpenAI ChatGPT and Anthropic Claude to assist with language editing, manuscript structure, LaTeX formatting, and reviewing the presentation of methodological details and results. The authors reviewed and edited the AI-assisted suggestions and take full responsibility for the accuracy and content of the manuscript.

\printbibliography[title={References}]

@article{cai2021,
  author  = {Yixin Cai and Fumiomi Takeda and Brian Foote and Lisa Wasko DeVetter},
  title   = {Effects of Machine-Harvest Interval on Fruit Quality of Fresh Market Northern Highbush Blueberry},
  journal = {Horticulturae},
  volume  = {7},
  number  = {8},
  pages   = {245},
  year    = {2021},
  doi     = {10.3390/horticulturae7080245}
}

@article{devetter2019,
  author  = {Lisa Wasko DeVetter and Wei Qiang Yang and Fumiomi Takeda and Scott Korthuis and Changying Li},
  title   = {Modified Over-the-Row Machine Harvesters to Improve Northern Highbush Blueberry Fresh Fruit Quality},
  journal = {Agriculture},
  volume  = {9},
  number  = {1},
  pages   = {13},
  year    = {2019},
  doi     = {10.3390/agriculture9010013}
}

@misc{ultralyticsPreprocessingDocs,
  author = {{Ultralytics}},
  title = {Ultralytics Data Augmentation API Reference: LetterBox},
  url = {https://docs.ultralytics.com/reference/data/augment/},
  urldate = {2026-11-19},
  note = {Online documentation}
}

@misc{openvinoToolkitDocs,
  author = {{OpenVINO}},
  title = {About OpenVINO},
  year = {2026},
  url = {https://docs.openvino.ai/2026/about-openvino.html},
  urldate = {2026-11-19},
  note = {Official documentation}
}

@article{sargent2020,
  author  = {Steven A. Sargent and Fumiomi Takeda and Jeffrey G. Williamson and Adrian D. Berry},
  title   = {Harvest of Southern Highbush Blueberry with a Modified, Over-The-Row Mechanical Harvester: Use of Handheld Shakers and Soft Catch Surfaces},
  journal = {Agriculture},
  volume  = {10},
  number  = {1},
  pages   = {4},
  year    = {2020},
  doi     = {10.3390/agriculture10010004}
}

@article{nuangjamnong2026,
  author  = {Nuangjamnong, Thanut and Duret, Steven and Flick, Denis and
             Moureh, Jean and Alvarez, Graciela},
  title   = {Modeling heat and mass transfer with water condensation process
             inside fruits bulk},
  journal = {Journal of Food Engineering},
  volume  = {407},
  pages   = {112831},
  year    = {2026},
  doi     = {10.1016/j.jfoodeng.2025.112831}
}

@article{sargent2021,
  author  = {Steven A. Sargent and Fumiomi Takeda and Jeffrey G. Williamson and Adrian D. Berry},
  title   = {Harvest of Southern Highbush Blueberry with a Modified, Over-the-Row Mechanical Harvester: Use of Soft-Catch Surfaces to Minimize Impact Bruising},
  journal = {Agronomy},
  volume  = {11},
  number  = {7},
  pages   = {1412},
  year    = {2021},
  doi     = {10.3390/agronomy11071412}
}

@article{jahari2015,
  author  = {Jahari, Mahirah and Yamamoto, Kazuya and Miyamoto, Munenori and
             Kondo, Naoshi and Ogawa, Yuichi and Suzuki, Tetsuhito and
             Habaragamuwa, Harshana and Ahmad, Usman},
  title   = {Double Lighting Machine Vision System to Monitor Harvested Paddy
             Grain Quality during Head-Feeding Combine Harvester Operation},
  journal = {Machines},
  year    = {2015},
  volume  = {3},
  number  = {4},
  pages   = {352--363},
  doi     = {10.3390/machines3040352}
}

@article{jahari2017,
  author  = {Jahari, Mahirah and Yamamoto, Kazuya and Miyamoto, Munenori and
             Kondo, Naoshi and Ogawa, Yuichi and Suzuki, Tetsuhito and
             Habaragamuwa, Harshana and Ahmad, Usman},
  title   = {Monitoring Harvested Paddy during Combine Harvesting Using a
             Machine Vision-Double Lighting System},
  journal = {Engineering in Agriculture, Environment and Food},
  year    = {2017},
  volume  = {10},
  number  = {2},
  pages   = {140--149},
  doi     = {10.1016/j.eaef.2016.12.001}
}

@article{chen2020,
  author  = {Jin Chen and Yi Lian and Rong Zou and Shuai Zhang and Xiaobo Ning and Mengna Han},
  title   = {Real-Time Grain Breakage Sensing for Rice Combine Harvesters Using Machine Vision Technology},
  journal = {International Journal of Agricultural and Biological Engineering},
  volume  = {13},
  number  = {3},
  pages   = {194--199},
  year    = {2020},
  doi     = {10.25165/j.ijabe.20201303.5478}
}

@article{wu2024,
  author  = {Zhiping Wu and Jin Chen and Zheng Ma and Yaoming Li and Yongle Zhu},
  title   = {Development of a Lightweight Online Detection System for Impurity Content and Broken Rate in Rice for Combine Harvesters},
  journal = {Computers and Electronics in Agriculture},
  volume  = {218},
  pages   = {108689},
  year    = {2024},
  doi     = {10.1016/j.compag.2024.108689}
}

@misc{cvat_software,
  author = {{CVAT.ai Corporation}},
  title = {{Computer Vision Annotation Tool (CVAT)}},
  year = {2023},
  howpublished = {Computer software},
  url = {https://github.com/cvat-ai/cvat},
  note = {Software}
}

@misc{jocher2026yolo26,
  title = {Ultralytics YOLO26: Unified Real-Time End-to-End Vision Models},
  author = {Jocher, Glenn and Qiu, Jing and Liu, Mengyu and Lyu, Shuai and Akyon, Fatih Cagatay and Kalfaoglu, Muhammet Esat},
  year = {2026},
  eprint = {2606.03748},
  archivePrefix = {arXiv},
  primaryClass = {cs.CV},
  doi = {10.48550/arXiv.2606.03748},
  url = {https://arxiv.org/abs/2606.03748}
}

@inproceedings{redmon2016yolo,
  author    = {Joseph Redmon and Santosh Divvala and Ross Girshick and Ali Farhadi},
  title     = {You Only Look Once: Unified, Real-Time Object Detection},
  booktitle = {Proceedings of the IEEE Conference on Computer Vision and Pattern Recognition},
  pages     = {779--788},
  year      = {2016}
}

@article{forney2022,
  author  = {Forney, Charles F. and Jordan, Michael A. and Pennell, Kathleen Munro and Fillmore, Sherry},
  title   = {Controlled atmosphere storage impacts fruit quality and flavor chemistry of five cultivars of highbush blueberry (Vaccinium corymbosum)},
  journal = {Postharvest Biology and Technology},
  volume  = {194},
  pages   = {112073},
  year    = {2022},
  doi     = {10.1016/j.postharvbio.2022.112073}
}

@article{pan2026,
  author  = {Yongfei Pan and Wenwen Guo and Jian Zhang and Minsheng Wu and Ang Zhao and Zhixi Deng and Ranbing Yang},
  title   = {An MCDE-YOLOv11-Based Online Detection Method for Broken and Impurity Rates in Potato Combine Harvesting},
  journal = {Agronomy},
  volume  = {16},
  number  = {7},
  pages   = {693},
  year    = {2026},
  doi     = {10.3390/agronomy16070693}
}

@article{li2014,
  author  = {Han Li and Won Suk Lee and Ku Wang},
  title   = {Identifying Blueberry Fruit of Different Growth Stages Using Natural Outdoor Color Images},
  journal = {Computers and Electronics in Agriculture},
  volume  = {106},
  pages   = {91--101},
  year    = {2014},
  doi     = {10.1016/j.compag.2014.05.015}
}

@article{yang2014,
  author  = {Ce Yang and Won Suk Lee and Paul Gader},
  title   = {Hyperspectral Band Selection for Detecting Different Blueberry Fruit Maturity Stages},
  journal = {Computers and Electronics in Agriculture},
  volume  = {109},
  pages   = {23--31},
  year    = {2014},
  doi     = {10.1016/j.compag.2014.08.009}
}

@article{esaki2025,
  author  = {Ikuma Esaki and Satoshi Noma and Takuya Ban and Rebeka Sultana and Ikuko Shimizu},
  title   = {Maturity Classification of Blueberry Fruit Using YOLO and Vision Transformer for Agricultural Assistance},
  journal = {Horticulturae},
  volume  = {11},
  number  = {10},
  pages   = {1272},
  year    = {2025},
  doi     = {10.3390/horticulturae11101272}
}

@article{liu2023,
  author  = {Yan Liu and Hongtao Zheng and Yonghua Zhang and Qiujie Zhang and Hongli Chen and Xueyong Xu and Gaoyang Wang},
  title   = {{``Is This Blueberry Ripe?''}: A Blueberry Ripeness Detection Algorithm for Use on Picking Robots},
  journal = {Frontiers in Plant Science},
  volume  = {14},
  pages   = {1198650},
  year    = {2023},
  doi     = {10.3389/fpls.2023.1198650}
}

@article{maceachern2023,
  author  = {Craig B. MacEachern and Travis J. Esau and Arnold W. Schumann and Patrick J. Hennessy and Qamar U. Zaman},
  title   = {Detection of Fruit Maturity Stage and Yield Estimation in Wild Blueberry Using Deep Learning Convolutional Neural Networks},
  journal = {Smart Agricultural Technology},
  volume  = {3},
  pages   = {100099},
  year    = {2023},
  doi     = {10.1016/j.atech.2022.100099}
}

@article{zhao2025,
  author  = {Kun Zhao and Yuhuan Li and Zunmin Liu},
  title   = {Three-Dimensional Spatial Perception of Blueberry Fruits Based on Improved YOLOv11 Network},
  journal = {Agronomy},
  volume  = {15},
  number  = {3},
  pages   = {535},
  year    = {2025},
  doi     = {10.3390/agronomy15030535}
}

@article{das2023,
  author  = {Anup Kumar Das and Travis J. Esau and Qamar U. Zaman and Aitazaz A. Farooque and Arnold W. Schumann and Patrick J. Hennessy},
  title   = {Machine Vision System for Real-Time Debris Detection on Mechanical Wild Blueberry Harvesters},
  journal = {Smart Agricultural Technology},
  volume  = {4},
  pages   = {100166},
  year    = {2023},
  doi     = {10.1016/j.atech.2022.100166}
}

@article{haydar2023,
  author  = {Zeeshan Haydar and Travis J. Esau and Aitazaz A. Farooque and Qamar U. Zaman and Patrick J. Hennessy and Kuljeet Singh and Farhat Abbas},
  title   = {Deep Learning Supported Machine Vision System to Precisely Automate the Wild Blueberry Harvester Header},
  journal = {Scientific Reports},
  volume  = {13},
  pages   = {10198},
  year    = {2023},
  doi     = {10.1038/s41598-023-37087-z}
}

@article{mullins2026,
  author  = {Connor C. Mullins and Travis J. Esau and Qamar U. Zaman and Ahmad A. Al-Mallahi and Aitazaz A. Farooque},
  title   = {Development of a Real-Time Non-Contact Yield Mapping System Using Cascading Neural Networks for Wild Blueberry Harvesters},
  journal = {Smart Agricultural Technology},
  volume  = {14},
  pages   = {102081},
  year    = {2026},
  doi     = {10.1016/j.atech.2026.102081}
}

@article{almallahi2010,
  author  = {Ahmad Al-Mallahi and Takashi Kataoka and Hiroshi Okamoto and Yoichi Shibata},
  title   = {Detection of Potato Tubers Using an Ultraviolet Imaging-Based Machine Vision System},
  journal = {Biosystems Engineering},
  volume  = {105},
  number  = {2},
  pages   = {257--265},
  year    = {2010},
  doi     = {10.1016/j.biosystemseng.2009.11.004}
}

@article{blok2025,
  author  = {Pieter M. Blok and Federico Magistri and Cyrill Stachniss and Haozhou Wang and James Burridge and Wei Guo},
  title   = {High-Throughput 3D Shape Completion of Potato Tubers on a Harvester},
  journal = {Computers and Electronics in Agriculture},
  volume  = {228},
  pages   = {109673},
  year    = {2025},
  doi     = {10.1016/j.compag.2024.109673}
}

@article{leiva2013,
  author  = {Gabriel A. Leiva-Valenzuela and Renfu Lu and Jos\'{e} Miguel Aguilera},
  title   = {Prediction of Firmness and Soluble Solids Content of Blueberries Using Hyperspectral Reflectance Imaging},
  journal = {Journal of Food Engineering},
  volume  = {115},
  number  = {1},
  pages   = {91--98},
  year    = {2013},
  doi     = {10.1016/j.jfoodeng.2012.10.001}
}

@article{fan2018,
  author  = {Shuxiang Fan and Changying Li and Wenqian Huang and Liping Chen},
  title   = {Data Fusion of Two Hyperspectral Imaging Systems with Complementary Spectral Sensing Ranges for Blueberry Bruising Detection},
  journal = {Sensors},
  volume  = {18},
  number  = {12},
  pages   = {4463},
  year    = {2018},
  doi     = {10.3390/s18124463}
}

@misc{liBlueberryDetectionDataset,
  author       = {Li, Zhengkun},
  title        = {Blueberry Detection Dataset},
  howpublished = {Kaggle dataset},
  url          = {https://www.kaggle.com/datasets/zhengkunli3969/blueberry-detection-dataset},
  urldate      = {2026-11-14}
}

@misc{liBlueberryMaturityDataset,
  author       = {Li, Zhengkun},
  title        = {Blueberry Maturity Classification},
  howpublished = {Kaggle dataset},
  url          = {https://www.kaggle.com/datasets/zhengkunli3969/blueberry-maturiety-classification},
  urldate      = {2026-11-14}
}

@inproceedings{lin2014coco,
  author    = {Lin, Tsung-Yi and Maire, Michael and Belongie, Serge J. and Hays, James and Perona, Pietro and Ramanan, Deva and Doll{\'a}r, Piotr and Zitnick, C. Lawrence},
  title     = {Microsoft {COCO}: Common Objects in Context},
  booktitle = {European Conference on Computer Vision},
  pages     = {740--755},
  year      = {2014},
  doi       = {10.1007/978-3-319-10602-1_48}
}

@misc{ultralyticsTrainDocs,
  author       = {{Ultralytics}},
  title        = {Model Training with Ultralytics {YOLO}},
  howpublished = {Online documentation},
  url          = {https://docs.ultralytics.com/modes/train/},
  urldate      = {2026-11-14}
}

@misc{ultralyticsAugmentDocs,
  author       = {{Ultralytics}},
  title        = {Data Augmentation using Ultralytics {YOLO}},
  howpublished = {Online documentation},
  url          = {https://docs.ultralytics.com/guides/yolo-data-augmentation/},
  urldate      = {2026-11-14}
}

@inproceedings{loshchilov2019decoupled,
  title={Decoupled Weight Decay Regularization},
  author={Ilya Loshchilov and Frank Hutter},
  booktitle={International Conference on Learning Representations (ICLR)},
  year={2019},
  url={https://openreview.net/forum?id=Bkg6RiCqY7}
}

\end{document}